\documentclass{article}
\usepackage{spconf,amsmath,graphicx}
\usepackage{booktabs}
\usepackage[utf8]{inputenc}
\usepackage{amsfonts}
\usepackage{hyperref}

\title{DisCoHead: Audio-and-Video-Driven Talking Head Generation by Disentangled Control of Head Pose and Facial Expressions}
\name{Geumbyeol Hwang\textsuperscript{*}, Sunwon Hong\textsuperscript{*}, Seunghyun Lee, Sungwoo Park, Gyeongsu Chae\textsuperscript{†}
\thanks{\textsuperscript{*}Equal contribution. \textsuperscript{†}Correspondence to: \href{mailto:gc@deepbrain.io}{\texttt{gc@deepbrain.io}} \newline
This work was supported by the Institute of Information \& Communications Technology Planning \& Evaluation (IITP) grant funded by the Ministry of Science and ICT (MSIT) of South Korea (No. 2021-0-00888).
}}
\address{DeepBrain AI Inc., Seoul, Korea}
\begin{document}
%\ninept
%
\maketitle
\begin{abstract}
For realistic talking head generation, creating natural head motion while maintaining accurate lip synchronization is essential. To fulfill this challenging task, we propose DisCoHead, a novel method to disentangle and control head pose and facial expressions without supervision. DisCoHead uses a single geometric transformation as a bottleneck to isolate and extract head motion from a head-driving video. Either an affine or a thin-plate spline transformation can be used and both work well as geometric bottlenecks. We enhance the efficiency of DisCoHead by integrating a dense motion estimator and the encoder of a generator which are originally separate modules. Taking a step further, we also propose a neural mix approach where dense motion is estimated and applied implicitly by the encoder. After applying the disentangled head motion to a source identity, DisCoHead controls the mouth region according to speech audio, and it blinks eyes and moves eyebrows following a separate driving video of the eye region, via the weight modulation of convolutional neural networks. The experiments using multiple datasets show that DisCoHead successfully generates realistic audio-and-video-driven talking heads and outperforms state-of-the-art methods. {\small Project page: }{\footnotesize \mbox{\url{https://deepbrainai-research.github.io/discohead/}}}
\end{abstract}
\begin{keywords}
Talking head generation, audio-and-video-driven, disentanglement, head pose, facial expressions
\end{keywords}
\section{Introduction}
\label{sec:intro}

Talking head generation is a method to synthesize facial video according to speech audio. It has broad applications like virtual videotelephony, automated video production, and character animation to name a few. Although deep learning technology has accelerated the advancement of talking head generation, creating natural head motion while maintaining accurate lip synchronization is still a challenge.

A way of talking face synthesis is inpainting the mouth parts in existing videos using audio information \cite{synthobama, wav2lip, nvp, audiodvp}. While head motion itself is natural in this approach, it is hard to manipulate the motion in existing videos, particularly to match the speech content.

Warping a facial frame based on a sequence of audio features is another way \cite{yousaidthat, song2019}. Speech and head motion are better aligned in this case, but the generated motions and nonspeech facial expressions lack natural dynamics. Following studies improve the audio-driven head motion and facial expressions by introducing sequence discriminators \cite{vougioukas2019, vougioukas2020} and a noise generator \cite{vougioukas2019}. \cite{yadav2020} adds diversity to the motion and expressions using a variational autoencoder framework. Other researches create head motion from audio via explicit structural representation of the face \cite{chen2020, makeittalk, zhang2021, audio2head}.

Few audio-driven talking face models control head pose across identities independent of lip synchronization. \cite{posecontrol} modularizes audio-visual representation into identity, head pose, and speech content using elaborate image augmentation and contrastive learning without structural intermediate information. Whereas \cite{apb2face, apb2facev2} use an external detector to extract shape-independent information of head pose and eye blinks, then fuse them with audio signals to reenact faces.

Meanwhile, in the field of image animation, a series of studies successfully decouple appearance and motion using unsupervised geometric-transformation-based approaches \cite{fomm, mraa, tpsmm}. In these approaches, the parameters of multiple affine or thin-plate spline (TPS) transformations between a source image and driving video frames are extracted, and dense motion (i.e., pixel-wise optical flow) is estimated based on the transformation parameters. Then, the source image is warped and inpainted to generate output frames.

\begin{figure*}
    \centering
    \includegraphics[width=17.8cm]{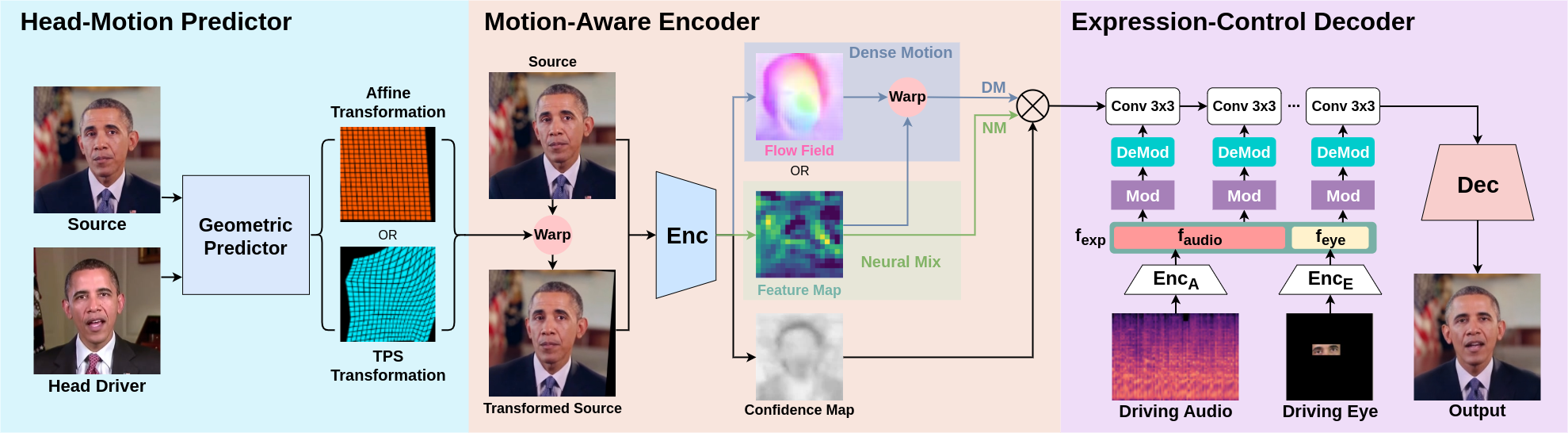}
    \caption{Architecture of our model. Head-Motion Predictor estimates an affine or TPS transformation by separately forwarding source and driving frames. With the original and transformed sources, Motion-Aware Encoder extracts head-motion-applied source features. Expression-Control Decoder then manipulates facial expressions according to driving audio and eye features.}
    \label{fig:fig1}
\end{figure*}

Inspired by the geometric-transformation-based image animation, we propose DisCoHead, a novel method enabling disentangled control of head pose and facial expressions for audio-and-video-driven talking head generation.

Our novelties are threefold. First, we use a single geometric transformation instead of multiple ones. A geometric transformation computed from a source and a driving frames serves as a bottleneck to isolate head pose from facial expressions. We find both affine and TPS transformations work well as geometric bottlenecks. Second, we integrate a dense motion estimator and the encoder of a generator into one. Unlike the existing image animation models, we do not need to combine multiple geometric transformations to estimate dense motion. With proper tweaks to input-output formulation, the encoder can effectively extract source features and dense motion simultaneously. Taking a step further to efficient architecture design, we also suggest a neural mix of the features from a source and a geometrically transformed source instead of explicit warping based on dense motion. Third, after warping or mixing the source features to apply head motion, we control facial expressions using speech audio and video frames of the eye region. We adopt the weight modulation of convolutional neural networks \cite{stylegan2}, where the audio and eye signals change the values of convolution filters to modify the facial expressions of the warped features.

Our proposed method allows us to separately control head pose, lip synchronization, and nonspeech facial expressions in an identity-agnostic manner. Our extensive experiments using multiple datasets show that DisCoHead successfully generates realistic audio-and-video-driven talking heads and outperforms state-of-the-art methods.

\section{Methods}
\label{sec:methods}

The entire pipeline of DisCoHead is depicted in Fig. \ref{fig:fig1}. We first introduce the head-motion predictor (Sec. \ref{ssec:predictor}). Then, we explain the motion-aware encoder (Sec. \ref{ssec:encoder}) and the expression-control decoder (Sec. \ref{ssec:decoder}) of the generator.

\subsection{Head-Motion Predictor}
\label{ssec:predictor}

The geometric transformation predictor representing head motion has two variants: affine predictor and TPS predictor.

We employ the PCA-based affine predictor \cite{mraa}. First, a single-channel heatmap $H$ is predicted from the input image $X$. The expectation of pixel locations $Z$ with probabilities $H$ can be interpreted as the translation component $t$ of an affine transformation:
\begin{equation}
    t=\sum_{z \in Z} H(z) z.
\end{equation}
The rotation and scaling components are computed with the principal axes of $H$, which can be extracted from a covariance matrix by the singular value decomposition (SVD):
\begin{equation}
    U \Sigma V=\sum_{z \in Z} H(z) (z - t) (z - t)^T.
\end{equation}
Here, $U$ and $V$ are unitary matrices, and $\Sigma$ is the diagonal matrix of singular values. Then we can configure the affine transformation as $A_{X \leftarrow R}=[U \Sigma^{1/2}, t]$, where $R$ is an abstract reference frame. The affine transformation for the motion from the head driver $D$ to the source $S$ is computed as:
\begin{equation}
    A_{S \leftarrow D} = A_{S \leftarrow R} A_{D \leftarrow R}^{-1}.
\end{equation}

For the TPS predictor \cite{tpsmm}, $N$ keypoints are predicted from the input frame. Given the $N$ pairs of keypoints from the source and the driving frames, we can get the corresponding TPS transformation as follows:
\begin{equation}
    T(p) = A
     \begin{bmatrix}
	  p \\
	  1 
	\end{bmatrix} 
	+ \sum_{i=1}^{N} \omega_i \phi(\lVert P_i^D - p \rVert_2),
\end{equation}
where $p = (x, y)^T$ is pixel coordinates, $P_i^D$ is the $i$th keypoint of the driving frame $D$, $A \in \mathbb{R}^{2\times3}$ and $\omega_i \in \mathbb{R}^{2\times1}$ are the TPS coefficients obtained by solving the minimum distortion equations in \cite{tps}, and a radial basis function $\phi(r)$ represents the influence of each keypoint on the pixel at $p$:
\begin{equation}
    \phi(r) = r^2 \log r^2.
\end{equation}

\subsection{Motion-Aware Encoder}
\label{ssec:encoder}

We integrate a dense motion estimator and the encoder of a generator into a motion-aware encoder that combines the source appearance with the driver's head pose and outputs a head-pose-aligned source feature to be used to generate the output frame. Along with the dense motion estimation approach (dense motion variant), we also propose a more compact approach that implicitly applies dense motion by neural networks (neural mix variant).

First, we warp the source frame $S$ with the predicted geometric transformation to make a transformed source $S_T$. The motion-aware encoder receives both $S$ and $S_T$ and produces a source feature $F$, a confidence map $C$, and optionally a motion mask $M$ for the dense motion variant:
\begin{equation}
    F, C, (M) = Enc(S, S_T).
\end{equation}

For the dense motion variant, the motion mask $M$ is used to produce a pixel-wise optical flow $O_P$ with the weighted sum of the identity flow $O_I$ and the coarse flow $O_T$ computed from the predicted geometric transformation. Then we warp $F$ with $O_P$ to make the aligned source feature $F_A$:
\begin{equation}\label{O_P}
    O_P = (1 - M) \circ O_I + M \circ O_T,
\end{equation}
\begin{equation}
    F_A = Warp(F, O_P).
\end{equation}

For the neural mix variant, we simply omit the warping based on the dense flow $O_P$ supposing that neural networks are able to implicitly estimate and apply dense motion by mixing the features from the source and the transformed source:
\begin{equation}
    F_{A} = F.
\end{equation}

The final encoder output $E$ is computed by applying the confidence map $C$ to the aligned feature $F_A$ to inform the decoder where and how much the local details should be inpainted: 
\begin{equation}\label{E}
    E = C \circ F_A.
\end{equation}

In Eq. (\ref{O_P}) and Eq. (\ref{E}), $\circ$ denotes the Hadamard product.

\subsection{Expression-Control Decoder}
\label{ssec:decoder}

An audio spectrogram corresponding to a target frame is encoded into an audio feature $f_{audio}$, and a masked eye driving frame is encoded into an eye feature $f_{eye}$. We concatenate the two features to obtain a facial expression feature $f_{exp}$. Then we manipulate the expressions of the encoder output $E$ by modulating the convolution kernel weights \cite{stylegan2} with $f_{exp}$:
\begin{equation}\label{eq11}
    \omega_{ijk}^{'}=\frac{s_i \cdot \omega_{ijk}}{\sqrt{\sum_{i,k} (s_i \cdot \omega_{ijk})^2 + \epsilon}},
\end{equation}
where $\omega$ and $\omega^{'}$ are the original and modulated weights, $s$ is the scale value predicted from $f_{exp}$, and $i$, $j$, and $k$ indicate the input channel, output channel, and spatial position of the convolution kernel, respectively. $\epsilon$ is a small constant to prevent numerical issues. 

After the modulated convolution blocks modify the facial expressions, DisCoHead generates the output frame with bilinear upsampling convolution blocks.

\begin{table*}[t]
\setlength{\tabcolsep}{0.67pt}
{\footnotesize
\begin{tabular}{l|cccccccccccccccccc}
\toprule
\multicolumn{1}{c|}{Dataset} & \multicolumn{6}{|c|}{Obama}& \multicolumn{6}{|c|}{GRID} & \multicolumn{6}{|c}{KoEBA} \\ \hline \hline
\multicolumn{1}{c|}{Method} & PSNR$\uparrow$ & SSIM$\uparrow$ & FID$\downarrow$ &  LPIPS$\downarrow$ & AKD$\downarrow$ & AED$\downarrow$
& \multicolumn{1}{|c}{PSNR$\uparrow$} & SSIM$\uparrow$ & FID$\downarrow$ &  LPIPS$\downarrow$ & AKD$\downarrow$ & AED$\downarrow$
& \multicolumn{1}{|c}{PSNR$\uparrow$} & SSIM$\uparrow$ & FID$\downarrow$ &  LPIPS$\downarrow$ & AKD$\downarrow$ & AED$\downarrow$ \\ \hline
APB2FaceV2 \cite{apb2facev2} & 17.84 & 0.494 & 60.38 & 0.596 & 3.746 & \multicolumn{1}{c|}{0.266}
                             & 28.12 & 0.662 & 62.24 & 0.261 & 2.952 & \multicolumn{1}{c|}{0.116}
                             & 22.04 & 0.706 & 58.59 & 0.256 & 2.972 & 0.244 \\
Wav2Lip \cite{wav2lip}       & 22.78 & 0.806 & 50.76 & 0.145 & 1.711 & \multicolumn{1}{c|}{0.075}
                             & 29.02 & 0.903 & 70.77 & 0.152 & 1.155 & \multicolumn{1}{c|}{0.024} 
                             & 25.61 & 0.876 & 32.85 & 0.086 & 1.834 & 0.080 \\
PC-AVS \cite{posecontrol}    & 22.14 & 0.491 & 8.493 & 0.227 & 2.761 & \multicolumn{1}{c|}{0.195} 
                             & 25.31 & 0.581 & 21.23 & 0.213 & 2.357 & \multicolumn{1}{c|}{0.102} 
                             & 23.36 & 0.556 & 37.45 & 0.101 & 2.174 & 0.128 \\ \hline
DisCoHead-Affine-DM & \textbf{28.39} & \textbf{0.904} & \textbf{0.618} & \textbf{0.051} &         0.915  & \multicolumn{1}{c|}{\textbf{0.031}} 
                    &         34.04  &         0.926  & \textbf{0.340} &         0.056  &         1.047  &         \multicolumn{1}{c|}{0.015} 
                    &         28.87  & \textbf{0.906} &         0.857  & \textbf{0.054} &         1.061  &                             0.023 \\

DisCoHead-Affine-NM &         27.07  &         0.883  &         0.736  &         0.052  &         1.002  &         \multicolumn{1}{c|}{0.037} 
                    & \textbf{34.14} & \textbf{0.928} &         0.406  &         0.076  &         1.007  & \multicolumn{1}{c|}{\textbf{0.012}} 
                    &         28.18  &         0.900  &         4.602  &         0.066  &         1.434  &                             0.043 \\

DisCoHead-TPS-DM    &         26.33  &         0.878  &         1.111  &         0.075  & \textbf{0.900} &         \multicolumn{1}{c|}{0.041} 
                    &         33.80  &         0.923  &         0.781  &         0.048  &         1.060  &         \multicolumn{1}{c|}{0.015} 
                    &         28.66  &         0.899  & \textbf{0.746} &         0.057  &         1.297  &                             0.033 \\

DisCoHead-TPS-NM    &         26.35  &         0.879  &         0.866  &         0.074  &         1.397           &\multicolumn{1}{c|}{0.062} 
                    &         33.81  &         0.924  &         0.849  & \textbf{0.046} & \textbf{0.983} & \multicolumn{1}{c|}{\textbf{0.012}} 
                    & \textbf{29.25} &         0.903  &         0.758  &         0.059  & \textbf{0.987} &                     \textbf{0.021} \\  
\bottomrule
\end{tabular}}
\caption{Quantitative results on the three datasets (DM: dense motion, NM: neural mix).}
\label{table:table1}
\end{table*}

\section{Experiments}
\label{sec:expts}

\subsection{Datasets}
\label{ssec:datasets}

We use the Obama dataset \cite{synthobama}, the GRID dataset \cite{grid}, and the Korean election broadcast addresses dataset (KoEBA). The Obama dataset contains the weekly presidential addresses of Barrack Obama. GRID is a set of video clips of 34 speakers pronouncing short utterances. KoEBA is a high-quality multi-speaker audio-video dataset composed of official broadcast addresses of Korean politicians.\footnote{\url{https://github.com/deepbrainai-research/koeba}} The facial regions of the video frames are cropped and resized to 256$\times$256. The driving audio is 400 ms long and centered on each video frame. We split each dataset into training (80\%) and test (20\%) sets.

\subsection{Implementation Details}
\label{ssec:impdets}

We adopt the hourglass network \cite{hourglass} for the affine predictor and ResNet-18 \cite{resnet} for the TPS predictor. The motion-aware encoder has three downsample convolution blocks, followed by three convolution layers with a kernel size of 1. The audio encoder consists of four 1D convolution layers, an LSTM layer, and two fully-connected layers. The eye encoder has five downsample blocks, followed by a global average pooling layer and two fully-connected layers. The expression-control decoder has six modulated residual convolution blocks, followed by three upsample convolution blocks and an output convolution layer with a kernel size of 7. We combine L1 loss and the perceptual loss \cite{percept} based on the pre-trained VGG-19 \cite{vgg} and use Adam optimizer with a learning rate of 1e-4 and a batch size of 16:
\begin{equation}\label{eq12}
L_{total} = L_{1} + \lambda \cdot L_{perceptual}.
\end{equation}

\subsection{Quantitative Results}

Table \ref{table:table1} shows the reconstruction performance of DisCoHead on the three datasets. We use peak signal-to-noise ratio (PSNR), structural similarity index measure (SSIM), Fréchet inception distance (FID), and learned perceptual image patch similarity (LPIPS) to assess generation quality. Average keypoint distance (AKD) measures the accuracy of pose and expressions using facial keypoints, and average Euclidean distance (AED) evaluates identity preservation based on a face recognition model. Comparisons with state-of-the-art baselines show that our model achieves the best results for all metrics and datasets. Both affine and TPS transformations work well as geometric bottlenecks to isolate and steer head motion. Also, the performance of the neural mix variant is almost on a par with the dense motion variant.

\subsection{Qualitative Results}

\begin{figure}[hb!]
    \centering
    \centerline{\includegraphics[width=6.9cm]{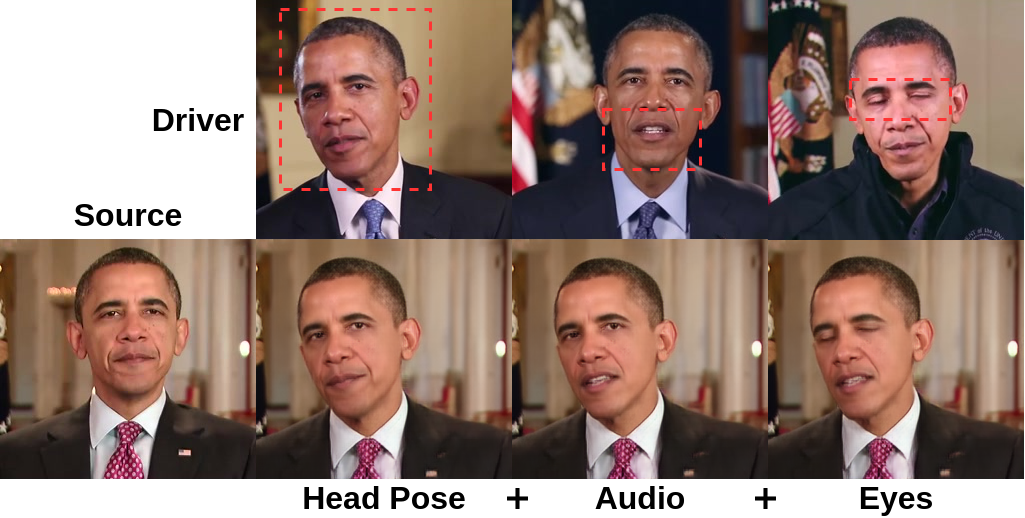}}
    \caption{Disentangled control of head pose, lip movements, and eye expressions on the Obama dataset.}
    \label{fig:fig2}
\end{figure}

\begin{figure}[hb!]
    \centering
    \includegraphics[width=5.5cm]{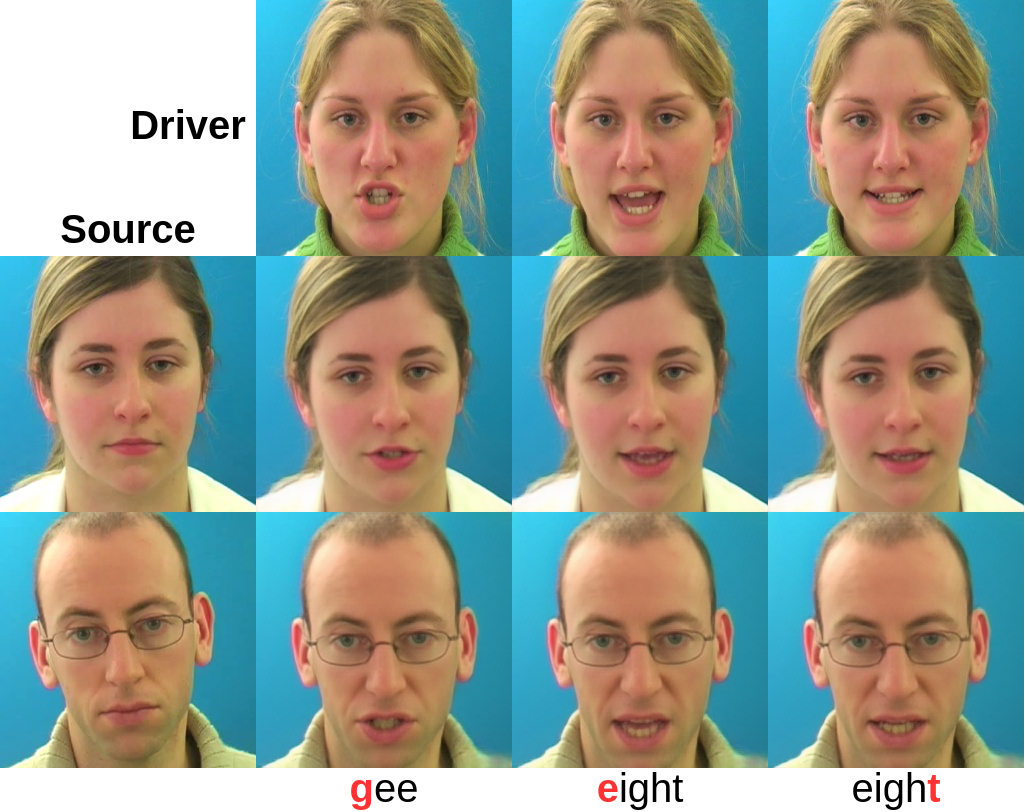}
    \caption{Qualitative results on the GRID dataset.}
    \label{fig:fig3}
\end{figure}

\begin{figure}[hb!]
    \centering
    \includegraphics[width=8.4cm]{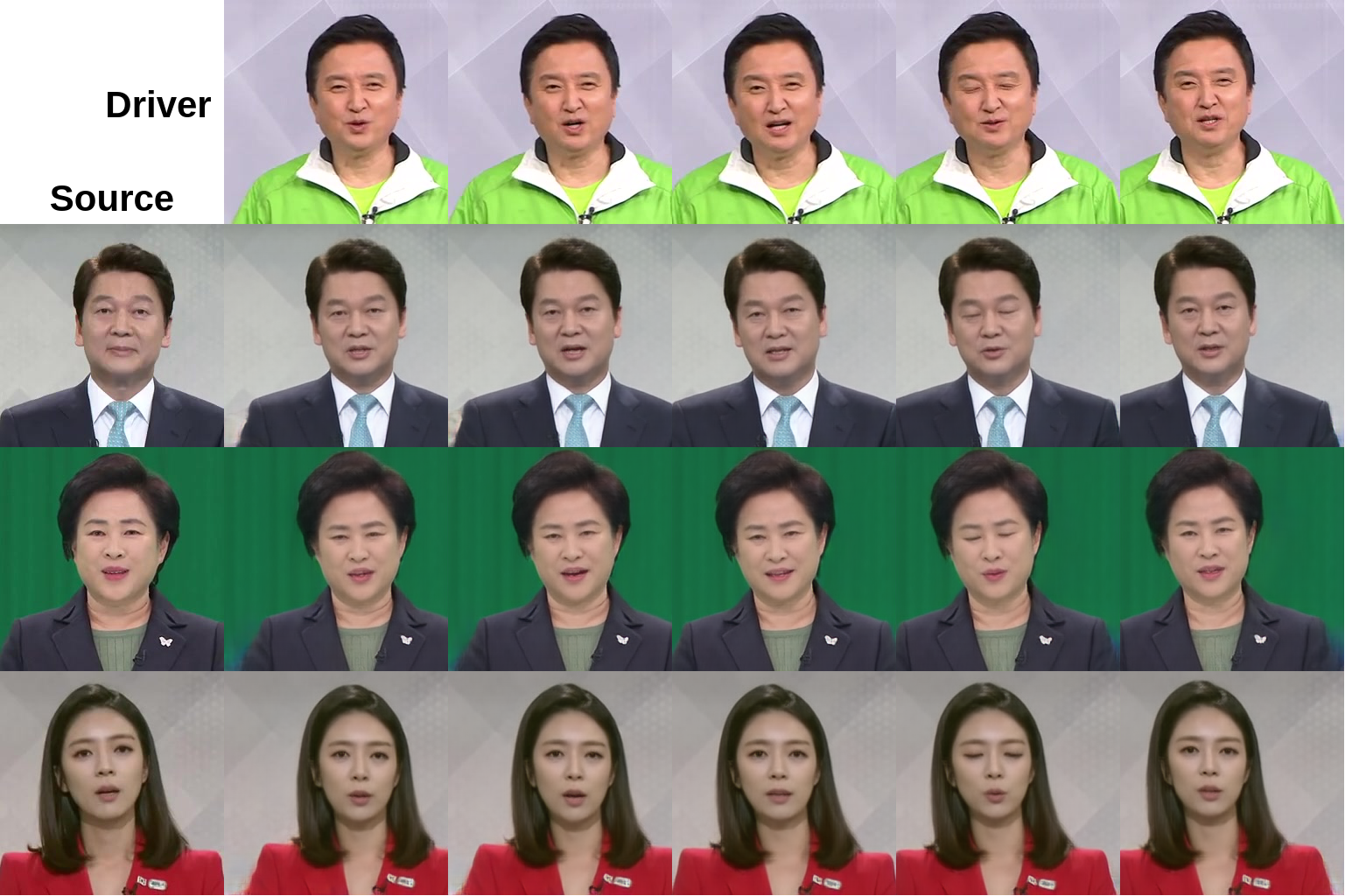}
    \caption{Qualitative results on the KoEBA dataset.}
    \label{fig:fig4}
\end{figure}

Fig. \ref{fig:fig2} demonstrates DisCoHead’s capability to disentangle head motion and facial expressions. The generated head pose, articulatory expression, and eye blink conform to the successively applied head pose, speech audio (corresponding to the shown driving frame), and eye blink of the excerpts from three different video clips without interfering with each other.

Fig. \ref{fig:fig3} and Fig. \ref{fig:fig4} present our model's ability to maintain the source identity on the GRID and KoEBA datasets. In Fig. \ref{fig:fig3}, the generated faces well maintain their own shapes, and the lip expressions pronouncing the same characters (marked in red) are correct but differ for each identity. In addition to precise motion control and identity preservation, subtle emotional expressions (i.e., slight frown) on the eye region of the driver are also transferred to the generated frames in Fig. \ref{fig:fig4}.

\section{Discussion}
\label{sec:disc}

DisCoHead's geometric bottleneck successfully disentangles identity, head pose, and facial expressions, but modeling extreme poses can be difficult. Its input audio and images of the eye region provide rich information to modulate expressions.

On the other hand, APB2FaceV2 \cite{apb2facev2} acquires head pose and eye blink information from an external facial landmark detector. Therefore, nonspeech facial expressions other than eye blinks cannot be delivered. Wav2Lip \cite{wav2lip} is unable to separate identity, head pose, and nonspeech expressions because it restores the lower half of a face according to the upper half and speech audio. Moreover, its output resolution is 96$\times$96, much smaller than 256$\times$256 of DisCoHead. PC-AVS \cite{posecontrol} can handle extreme poses but only decomposes identity, head pose, and speech content. As a consequence, it lacks the ability to control nonspeech facial expressions.

\section{Conclusion}
\label{sec:con}

We design a novel method to disentangle and separately control head pose and facial expressions for audio-and-video-driven talking head generation. DisCoHead uses a single geometric transformation as a medium to represent head motion and the weight modulation of convolutional layers to manipulate speech and nonspeech facial expressions. We enhance the efficiency of DisCoHead by fusing dense motion estimation and video frame generation which are originally formulated as a sequential process. In our experiments using multiple datasets, DisCoHead excels state-of-the-art methods.

% \vfill\pagebreak

% \section{REFERENCES}
% \label{sec:refs}

\bibliographystyle{IEEEbib}
{\small
\bibliography{refs}

\begin{thebibliography}{10}

\bibitem{synthobama}
Supasorn Suwajanakorn, Steven~M. Seitz, and Ira Kemelmacher-Shlizerman,
\newblock ``Synthesizing obama: Learning lip sync from audio,''
\newblock {\em ACM Trans. Graph.}, vol. 36, no. 4, jul 2017.

\bibitem{wav2lip}
K~R Prajwal, Rudrabha Mukhopadhyay, Vinay~P. Namboodiri, and C.V. Jawahar,
\newblock ``A lip sync expert is all you need for speech to lip generation in
  the wild,''
\newblock in {\em Proceedings of the 28th ACM International Conference on
  Multimedia}, 2020, p. 484–492.

\bibitem{nvp}
Justus Thies, Mohamed Elgharib, Ayush Tewari, Christian Theobalt, and Matthias
  Nie{\ss}ner,
\newblock ``Neural voice puppetry: Audio-driven facial reenactment,''
\newblock in {\em Computer Vision -- ECCV 2020}, 2020, pp. 716--731.

\bibitem{audiodvp}
Xin Wen, Miao Wang, Christian Richardt, Ze-Yin Chen, and Shi-Min Hu,
\newblock ``Photorealistic audio-driven video portraits,''
\newblock {\em IEEE Transactions on Visualization and Computer Graphics}, vol.
  26, no. 12, pp. 3457--3466, 2020.

\bibitem{yousaidthat}
Joon~Son Chung, Amir Jamaludin, and Andrew Zisserman,
\newblock ``You said that?,''
\newblock in {\em Proceedings of the British Machine Vision Conference (BMVC)},
  September 2017, pp. 109.1--109.12.

\bibitem{song2019}
Yang Song, Jingwen Zhu, Dawei Li, Andy Wang, and Hairong Qi,
\newblock ``Talking face generation by conditional recurrent adversarial
  network,''
\newblock in {\em Proceedings of the Twenty-Eighth International Joint
  Conference on Artificial Intelligence, {IJCAI-19}}, 7 2019, pp. 919--925.

\bibitem{vougioukas2019}
Konstantinos Vougioukas, Stavros Petridis, and Maja Pantic,
\newblock ``End-to-end speech-driven realistic facial animation with temporal
  gans,''
\newblock in {\em Proceedings of the IEEE/CVF Conference on Computer Vision and
  Pattern Recognition (CVPR) Workshops}, June 2019.

\bibitem{vougioukas2020}
Konstantinos Vougioukas, Stavros Petridis, and Maja Pantic,
\newblock ``Realistic speech-driven facial animation with gans,''
\newblock {\em Int. J. Comput. Vision}, vol. 128, no. 5, pp. 1398–1413, may
  2020.

\bibitem{yadav2020}
Ravindra Yadav, Ashish Sardana, Vinay~P. Namboodiri, and Rajesh~M. Hegde,
\newblock ``Stochastic talking face generation using latent distribution
  matching,''
\newblock in {\em Proc. Interspeech 2020}, 2020, pp. 1311--1315.

\bibitem{chen2020}
Lele Chen, Guofeng Cui, Celong Liu, Zhong Li, Ziyi Kou, Yi~Xu, and Chenliang
  Xu,
\newblock ``Talking-head generation with rhythmic head motion,''
\newblock in {\em Computer Vision -- ECCV 2020}, 2020, pp. 35--51.

\bibitem{makeittalk}
Yang Zhou, Xintong Han, Eli Shechtman, Jose Echevarria, Evangelos Kalogerakis,
  and Dingzeyu Li,
\newblock ``Makelttalk: Speaker-aware talking-head animation,''
\newblock {\em ACM Trans. Graph.}, vol. 39, no. 6, nov 2020.

\bibitem{zhang2021}
Zhimeng Zhang, Lincheng Li, Yu~Ding, and Changjie Fan,
\newblock ``Flow-guided one-shot talking face generation with a high-resolution
  audio-visual dataset,''
\newblock in {\em Proceedings of the IEEE/CVF Conference on Computer Vision and
  Pattern Recognition (CVPR)}, June 2021, pp. 3661--3670.

\bibitem{audio2head}
Suzhen Wang, Lincheng Li, Yu~Ding, Changjie Fan, and Xin Yu,
\newblock ``Audio2head: Audio-driven one-shot talking-head generation with
  natural head motion,''
\newblock in {\em Proceedings of the Thirtieth International Joint Conference
  on Artificial Intelligence, {IJCAI-21}}, 8 2021, pp. 1098--1105.

\bibitem{posecontrol}
Hang Zhou, Yasheng Sun, Wayne Wu, Chen~Change Loy, Xiaogang Wang, and Ziwei
  Liu,
\newblock ``Pose-controllable talking face generation by implicitly modularized
  audio-visual representation,''
\newblock in {\em Proceedings of the IEEE/CVF Conference on Computer Vision and
  Pattern Recognition (CVPR)}, June 2021, pp. 4176--4186.

\bibitem{apb2face}
Jiangning Zhang, Liang Liu, Zhucun Xue, and Yong Liu,
\newblock ``Apb2face: Audio-guided face reenactment with auxiliary pose and
  blink signals,''
\newblock in {\em ICASSP 2020 - 2020 IEEE International Conference on
  Acoustics, Speech and Signal Processing (ICASSP)}, 2020, pp. 4402--4406.

\bibitem{apb2facev2}
Jiangning Zhang, Xianfang Zeng, Chao Xu, and Yong Liu,
\newblock ``Real-time audio-guided multi-face reenactment,''
\newblock {\em IEEE Signal Processing Letters}, vol. 29, pp. 1--5, 2022.

\bibitem{fomm}
Aliaksandr Siarohin, St\'{e}phane Lathuili\`{e}re, Sergey Tulyakov, Elisa
  Ricci, and Nicu Sebe,
\newblock ``First order motion model for image animation,''
\newblock in {\em Advances in Neural Information Processing Systems}, 2019,
  vol.~32.

\bibitem{mraa}
Aliaksandr Siarohin, Oliver~J. Woodford, Jian Ren, Menglei Chai, and Sergey
  Tulyakov,
\newblock ``Motion representations for articulated animation,''
\newblock in {\em Proceedings of the IEEE/CVF Conference on Computer Vision and
  Pattern Recognition (CVPR)}, June 2021, pp. 13653--13662.

\bibitem{tpsmm}
Jian Zhao and Hui Zhang,
\newblock ``Thin-plate spline motion model for image animation,''
\newblock in {\em Proceedings of the IEEE/CVF Conference on Computer Vision and
  Pattern Recognition (CVPR)}, June 2022, pp. 3657--3666.

\bibitem{stylegan2}
Tero Karras, Samuli Laine, Miika Aittala, Janne Hellsten, Jaakko Lehtinen, and
  Timo Aila,
\newblock ``Analyzing and improving the image quality of stylegan,''
\newblock in {\em Proceedings of the IEEE/CVF Conference on Computer Vision and
  Pattern Recognition (CVPR)}, June 2020.

\bibitem{tps}
F.L. Bookstein,
\newblock ``Principal warps: thin-plate splines and the decomposition of
  deformations,''
\newblock {\em IEEE Transactions on Pattern Analysis and Machine Intelligence},
  vol. 11, no. 6, pp. 567--585, 1989.

\bibitem{grid}
Martin Cooke, Jon Barker, Stuart Cunningham, and Xu~Shao,
\newblock ``An audio-visual corpus for speech perception and automatic speech
  recognition,''
\newblock {\em The Journal of the Acoustical Society of America}, vol. 120, pp.
  2421—2424, November 2006.

\bibitem{hourglass}
Alejandro Newell, Kaiyu Yang, and Jia Deng,
\newblock ``Stacked hourglass networks for human pose estimation,''
\newblock in {\em Computer Vision -- ECCV 2016}, 2016, pp. 483--499.

\bibitem{resnet}
Kaiming He, Xiangyu Zhang, Shaoqing Ren, and Jian Sun,
\newblock ``Deep residual learning for image recognition,''
\newblock in {\em Proceedings of the IEEE Conference on Computer Vision and
  Pattern Recognition (CVPR)}, June 2016.

\bibitem{percept}
Justin Johnson, Alexandre Alahi, and Li~Fei-Fei,
\newblock ``Perceptual losses for real-time style transfer and
  super-resolution,''
\newblock in {\em Computer Vision -- ECCV 2016}, 2016, pp. 694--711.

\bibitem{vgg}
Karen Simonyan and Andrew Zisserman,
\newblock ``Very deep convolutional networks for large-scale image
  recognition,''
\newblock in {\em International Conference on Learning Representations}, 2015.

\end{thebibliography}
}
\end{document}